\documentclass[10pt,twocolumn,letterpaper]{article}

\usepackage{cvpr}              % To produce the CAMERA-READY version

\usepackage{graphicx}
\usepackage{amsmath}
\usepackage{amssymb}
\usepackage{booktabs}
\usepackage{times}
\usepackage{epsfig}
\usepackage{graphicx}

\usepackage{algorithm}
\usepackage{algorithmic}
\renewcommand{\algorithmicrequire}{\textbf{Input:}}
\renewcommand{\algorithmicensure}{\textbf{Output:}}

\usepackage[pagebackref,breaklinks,colorlinks]{hyperref}

\usepackage[capitalize]{cleveref}
\crefname{section}{Sec.}{Secs.}
\Crefname{section}{Section}{Sections}
\Crefname{table}{Table}{Tables}
\crefname{table}{Tab.}{Tabs.}

\def\confName{CVPR}
\def\confYear{2022}

\makeatletter
\renewcommand{\@fnsymbol}[1]{%
  \ifcase#1\or †\or *\or §\or ‖\or ‡\or **\else\@ctrerr\fi}
\makeatother

\title{MESC-3D:Mining Effective Semantic Cues for 3D Reconstruction \\from a Single Image}
\author{Shaoming Li\textsuperscript{1}\thanks{Equal contribution.}, 
Qing Cai\textsuperscript{1†}\thanks{The corresponding author is Qing Cai.},
Songqi Kong$^1$, Runqing Tan$^1$, Heng Tong$^1$,\\ Shiji Qiu$^1$, Yongguo Jiang$^1$, Zhi Liu$^2$\\
Faculty of Computer Science and Technology, Ocean University of China$^1$\\
School of Information Science and Engineering, Shandong University$^2$\\
{\tt\small \{lishaoming,kongsongqi,tanrunqing,tongheng,qiushiji\}@stu.ouc.edu.cn}\\
{\tt\small \{cq,jiangyongguo\}@ouc.edu.cn\hspace {3mm}liuzhi@sdu.edu.cn}
}

\let\oldtwocolumn\twocolumn
\renewcommand\twocolumn[1][]{%
    \oldtwocolumn[{#1}{
    \begin{center}
           \includegraphics[width=\textwidth]{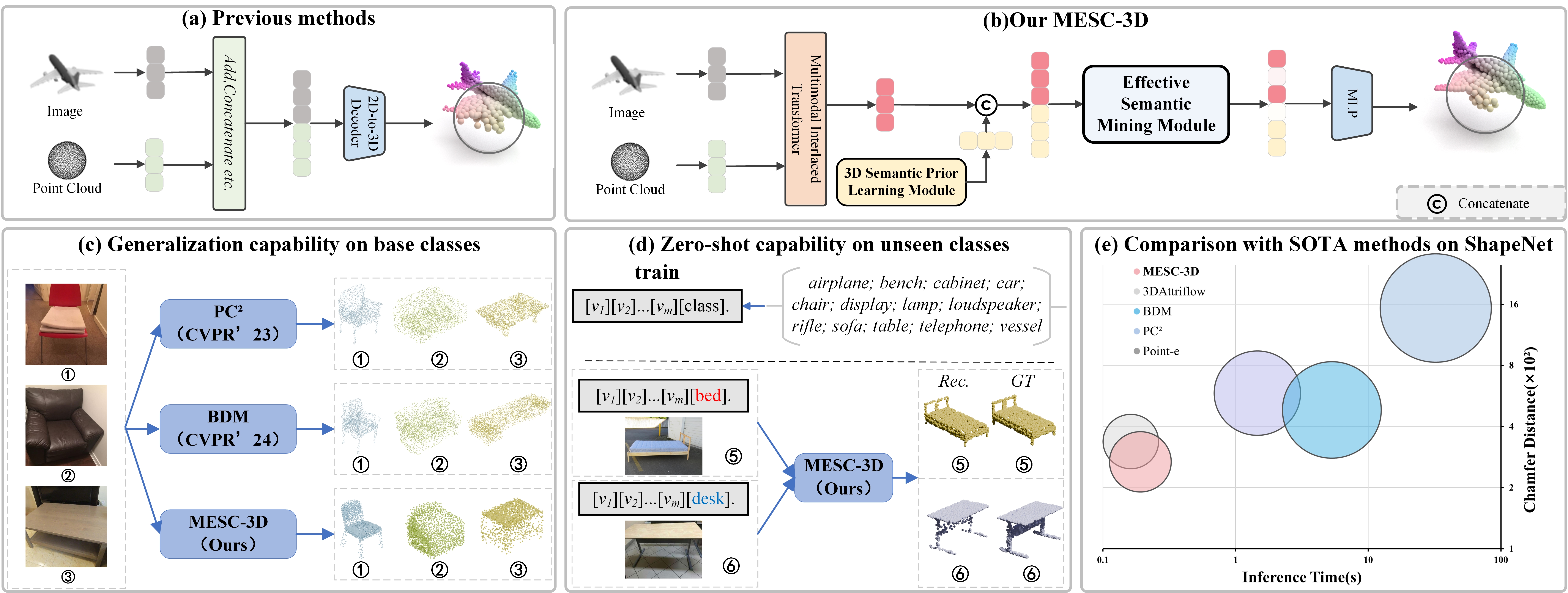}
           \captionof{figure}{ (a) Previous methods simply performed basic operations on the extracted 2D image information and 3D point cloud without establishing a connection between them. 
 (b) Compared to that, we introduced two key designs: First, the Effective Semantic Mining Module, which effectively mines semantic information from the entangled features and enables point cloud to select the information. Second, the 3D Semantic Prior Learning Module, which aims to enable the model to interpret 3D structures as humans do in 3D reconstruction from a single image. (c) The generalization comparsion between the proposed MESC-3D and SOTA methods on base classes with complex backgrounds. (d) MESC-3D's zero-shot on unseen classes. (e) Comparison with state-of-the-art methods on ShapeNet \cite{chang2015shapenet} Dataset on Chamfer Distance (y-axis), parameter count (size of the area), and inference time (x-aixs) which show that MESC-3D achieve the best performance with comparable computational cost. }
           \label{introduction}
        \end{center}
    }]}
\begin{document}
\maketitle

\begin{abstract}
Reconstructing 3D shapes from a single image plays an important role in computer vision. Many methods have been proposed and achieve impressive performance. However, existing methods mainly focus on extracting semantic information from images and then simply concatenating it with 3D point clouds without further exploring the concatenated semantics. As a result, these entangled semantic features significantly hinder the reconstruction performance. In this paper, we propose a novel single-image 3D reconstruction method called Mining Effective Semantic Cues for 3D Reconstruction from a Single Image (MESC-3D), which can actively mine effective semantic cues from entangled features. Specifically, we design an Effective Semantic Mining Module to establish connections between point clouds and image semantic attributes, enabling the point clouds to autonomously select the necessary information. Furthermore, to address the potential insufficiencies in semantic information from a single image, such as occlusions, inspired by the human ability to represent 3D objects using prior knowledge drawn from daily experiences, we introduce a 3D Semantic Prior Learning Module. This module incorporates semantic understanding of spatial structures, enabling the model to interpret and reconstruct 3D objects with greater accuracy and realism, closely mirroring human perception of complex 3D environments. Extensive evaluations show that our method achieves significant improvements in reconstruction quality and robustness compared to prior works. Additionally, further experiments validate the strong generalization capabilities and excels in zero-shot preformance on unseen classes. Code is available at \url{https://github.com/QINGQINGLE/MESC-3D}. 

\end{abstract}  
\section{Introduction}
Image-based 3D object reconstruction plays an important role in computer vision \cite{melas2023realfusion,pointe,3Dcomputer,sketch,pc2}, with widespread applications in areas such as virtual reality \cite{VR,VR_2,VR_3}, autonomous driving \cite{autonomous_driving,autonomous_driving_2,autonomous_driving_3}, video games, robotics \cite{robotic} and 3D content creation \cite{yang2023asm}. So far, many image-based 3D object reconstruction method have been proposed, which can be broadly categorized into single-view 3D shape reconstruction \cite{3dAttriflow,bdm} and multi-view 3D shape reconstruction \cite{wang2021multi,zhou2023sparsefusion}. This paper focuses on single-view point cloud reconstruction and conducts a series of studies on this topic. The typical paradigm for single-view 3D point cloud reconstruction involves extracting semantic attributes from a single image, obtaining implicit or explicit features through these attributes, parsing the image’s class and geometric information, and finally decoding it into a 3D shape via a 3D decoder. In this process, understanding the semantic attributes in the image is crucial \cite{crevier1997knowledge,liu2024comprehensive} and serves as the first step in performing reconstruction tasks \cite{xue2023ulip,xue2024ulip2,zhang2022pointclip}. The typical paradigm followed by most existing methods \cite{3dAttriflow,xie2019pix2vox,xie2020pix2vox++,pixel2mesh,pixel2mesh++} indicates that the key to 2D-to-3D reconstruction lies in accurately translating the semantic attributes of the image into 3D space.

However, most existing methods \cite{3dAttriflow,pixel2mesh,pixel2mesh++,xie2019pix2vox,xie2020pix2vox++} simply transfer image semantic information to the 3D decoder by performing element-wise addition or channel concatenation between semantic features and the 3D point cloud as seen in Fig. \ref{introduction}(a). Since these semantic features are entangled, directly using them for reconstruction would severely affect the reconstruction quality. Moreover, due to the limited visual information in a single view, 3D reconstruction from a single image often produce rough geometric shapes in occluded regions. Despite the difficulty of this task, humans are adept at using a range of monocular cues and prior knowledge to infer the 3D structure of common objects from single view. Inspired by daily life, humans can easily and naturally infer the shape of the back of an object in a photograph based on life experiences. 

Based on the above discussion, this paper aims to design a novel network that, like humans, can sketch the shape of an object in theirs mind based on prior understanding when observing the object in an image. To achieve this goal, this paper has made two key designs, as shown in Fig. \ref{introduction} (b). Specifically, to simulate the process of humans obtaining prior knowledge of the object when seeing it in a single image, we design a 3D Semantic Prior Learning Module (3DSPL), which integrates object global semantic prior into the network by utilizing a learnable text prompt. Furthermore, to further untangle semantics, we have designed an Effective Semantic Mining Module (ESM), which can enable point cloud to autonomously select the necessary semantic information and establish a connection between point cloud and semantic attributes. Experimental results also demonstrate the effectiveness of this design.

To summarize, our contributions are:
\begin{itemize}
\item We design an Effective Semantic Mining Module (ESM). Unlike traditional 3D reconstruction\cite{3dAttriflow,xie2019pix2vox,xie2020pix2vox++,pixel2mesh,pixel2mesh++}  that perform simple feature operations between the image's semantic attributes and the point cloud, our approach allows the point cloud to autonomously select semantic information, thereby establishing a connection between the point cloud and the features.
\item We design a 3D Semantic Prior Learning Module (3DSPL). Through contrastive learning between text and point cloud modalities, this module effectively compensates for the lack of semantic information in single view, enabling it to infer object shapes from a single image, much like how humans can intuitively deduce the complete structure of an object from a single image.
\item We conduct extensive experiments to demonstrate the versatile capabilities of MESC-3D. It exhibits superior performance on synthetic dataset \cite{chang2015shapenet} and real-world dataset \cite{sun2018pix3d}, generalization capability and zero-shot capability on 3D reconstruction. 
\end{itemize}

\begin{figure*}[!t]
    \centering
    \includegraphics[width=\linewidth]{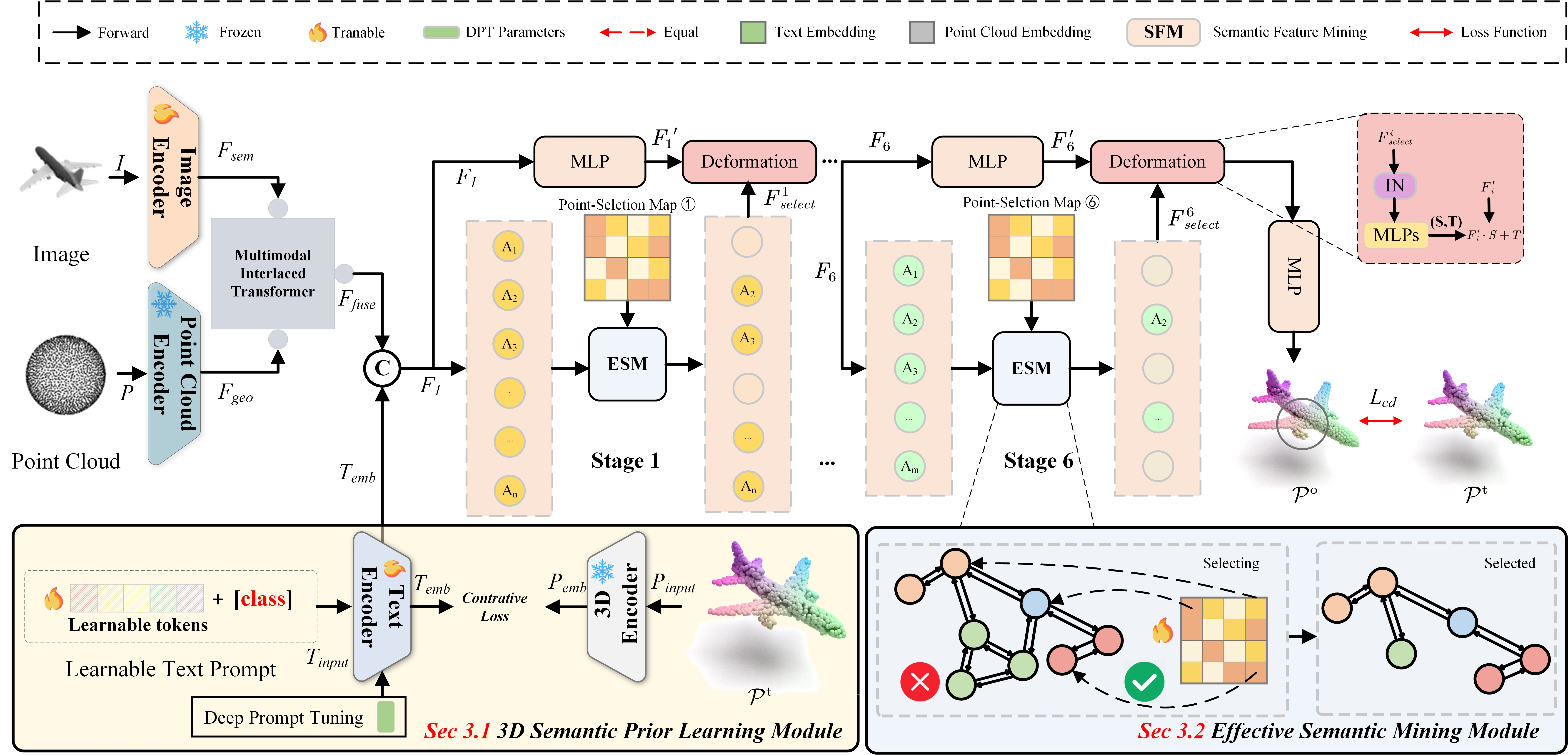} 
    \caption{\textbf{The overall architecture of MESC-3D.} Our network is composed of two main components. (a) The 3DSPL align point cloud modality features with text features, aiming to capture the unique 3D geometric characteristics of each category. (b) The ESM establishes a connection between the semantic feature $F_{i}$ and the 3D point cloud at $i^{th}$ stage, allowing each point to autonomously select the most valuable semantic information.}
    \label{pipeline}
\end{figure*}

\section{Related Work}
\textbf{Single-View Point Clouds Reconstruction.} 3D reconstruction is a fundamental task of translating the 2D world into a 3D representation. Unlike meshes and voxels, point clouds offer a memory-efficient way to represent the shape of arbitrary objects. Point cloud reconstruction methods are currently divided into traditional deep learning-based approaches \cite{3dr2n2,3dAttriflow,pixel2mesh,xie2019pix2vox} and diffusion-based methods \cite{pointe,dit3d,pc2}. Traditional deep learning methods include approaches like 3DR2N2 \cite{3dr2n2}, PSGN \cite{psgn}, Pix2Mesh \cite{pixel2mesh}, AtlasNet \cite{AtlasNet}, OccNet \cite{OCcnet} and 3DAttriflow \cite{3dAttriflow}. These methods typically involve extracting semantic attribute from single or multiple views of images and then decoding these 2D features into 3D space. They have demonstrated effective performance in single-view and multi-view reconstruction \cite{wang2021multi}. In recent years, researchers \cite{pointe,dit3d,pc2,cui2024lam3d} have explored various pipelines based on diffusion probabilistic models to achieve 3D shape generation. For instance, methods such as Point-E \cite{pointe}, PC² \cite{pc2} and BDM \cite{bdm} have successfully generated sparse point clouds from single images through a denoising diffusion process. They use a diffusion model which conditions on image. Additionally, PC² takes its camera pose and gradually denoises a set of 3D points, aiming to project local image features onto the partially-denoised point cloud at each step of the diffusion process. Unfortunately, these methods \cite{3dAttriflow,pixel2mesh,pixel2mesh++,xie2019pix2vox,xie2020pix2vox++} either focus on enhancing the richness of semantic information or improving communication between 2D and 3D encoders, without adequately addressing the connection between these modalities. As a result, these methods \cite{3dAttriflow,pixel2mesh,pixel2mesh++,xie2019pix2vox,xie2020pix2vox++} fail to fully exploit the effective semantic information in the image. Furthermore, a single view image often suffer from insufficient semantic information primarily due to occlusion.

\noindent\textbf{Prompt Learning.} This topic \cite{chen2022prompt,derakhshani2023bayesianprompt,cocoop,zhou2022learningprompt,wu2024vadclip} has emerged as an effective fine-tuning strategy to adapt Vision models. It adds a small number of learnable embeddings along with model inputs which are optimized during training while the rest of the model is kept frozen. This approach has already shown outstanding performance in the 2D domain. It is worth noting that the CoOp \cite{zhou2022learningprompt} early presents the first comprehensive study about adapting large vision models with prompt learning. Specifically, CoOp \cite{zhou2022learningprompt} models a prompt’s context words with learnable vectors while the entire pre-trained parameters are kept fixed. It demonstrates that prompt learning could bring much stronger robustness. Following this, they introduced CoCoOp \cite{cocoop}. Inspired by their work, we aim to extend prompt learning to the 3D domain, addressing prompt engineering for 3D and learning prior of 3D structure like humans, making use of the 3D rich knowledge encoded in the parameters to optimize the context. We attempt to incorporate 3D geometric prior into the Text Encoder to enhance guidance for subsequent 3D reconstruction tasks.

\section{Method}
The overall architecture of MESC-3D is visualized in Fig. \ref{pipeline}. Given a single view Image $I\in[224,224]$ and a random sphere points $P\in(2048,3)$. We first adopt ResNet18 \cite{resnet18} and pre-trained PointMAE \cite{pointmae} to obtain the image feature $F_{sem}$ and the point cloud feature $F_{geo}$ respectively. Then, using the Multimodal Interlaced Transformer $MIT$ \cite{interlaced}, we enrich these features through mutual interaction, resulting in the fused feature $F_{fuse} = MIT(F_{geo}, F_{sem})$. Next, a prior feature $T_{emb}$ obtained by one of the key component 3DSPL, which is then concatenated with the original feature to form the stage 1 feature as $F_{1}={Concat}[F_{fuse},T_{emb}]$. Finally, after six stages of effective feature selection by using another main component ESM, the 3D coordinates of each point are predicted by a tiny MLP.

\subsection{3D Semantic Prior Learning Module}
\label{sec:text encoder}
In order to endow the model with the ability to infer like humans when seeing an object, we utilize contrastive learning to imitate the human acquisition of prior knowledge in daily life. This enables the model to recognize an object's category and associate it with learned prior knowledge for accurate 3D reconstruction. We use the ShapeNet \cite{chang2015shapenet} dataset as our 3D modality, which is one of the most extensive public 3D CAD datasets. However, instead of using these specific textual descriptions, we replace them with learnable text prompt.

\noindent\textbf{Deep Prompt Tuning(DPT)} By using Deep Prompt Tuning (DPT) \cite{zhou2023zegclip}, we can retain the original CLIP parameters intact while effectively learning 3D geometric prior. Formally, we denote the original text embeddings in CLIP as $K^l=\{k_1^l,k_2^l,\ldots,k_N^l\}$. Deep prompt tuning appends learnable tokens $P^l=\{p_1^l,p_2^l,\ldots,p_M^l\}$ to this token sequence in each Vision Transformer (ViT) layer of the CLIP text encoder. Then, the $\text{l}$-th Multi-Head Attention (MHA) module processes the input token as:
\begin{equation}
    [_-,K^l]=\text{Layer}^l[P^{l-1},K^{l-1}],
\end{equation}
where the output embeddings of $\{p_1^l,p_2^l,\ldots,p_M^l\}$ are discarded (denoted as $_-$) and are not fed into the next layer. Therefore, $\{p_1^l,p_2^l,\ldots,p_M^l\}$ merely acts as a set of learnable parameters to adapt the MHA model.

 \noindent\textbf{Learnable Text Prompt} Inspired by CoOp \cite{zhou2022learningprompt}, we designed a learnable text prompt specifically for the 3D reconstruction task. To extract useful information for 3D reconstruction, we allow the model to autonomously summarize the corresponding 3D textual descriptions for point cloud. The learnable text prompt shares the same context vectors with all classes. The design of continuous representations also allows full exploration in the word embedding space, which facilitates the learning of 3D-relevant context. In this process, we only need to provide an appropriate prompt, such as the 3D shape category. Concretely, the original 3D categories are first transformed into class tokens through the CLIP tokenizer, i.e., $t_{\mathrm{init}}=\mathrm{Tokenizer(Category)}$, where Category is the discrete 3D text category, e.g., airplane, vessel, etc. We then concatenate $t_{\mathrm{init}}$ with the learnable text prompt $\{c_1,\ldots,c_l\}$, which contains $\text{l}$ context tokens, to form a complete sentence token. Thus, the input to the text encoder is presented as follows:
\begin{equation}
    t_p=\{c_1,...,t_{init},...,c_l\}.
\end{equation}

Here, we place the class token at the middle of a sequence. Then this sequence token is added to the positional embedding to obtain positional information, and finally, the text encoder of CLIP takes as input $T_{\mathrm{input}}$ and generates prompt embedding $T_{\mathrm{emb}}$.

\subsection{Effective Semantic Mining Module}
\label{sec:Semantic mining}
The model acquires 3D prior knowledge $T_{emb}$ through the 3DSPL. This prior knowledge is then integrated with enriched semantic features $F_{fuse}$, resulting in a more comprehensive 3D attribute representation $F_{1}$. Through multi-stage selection, the model progressively refines attribute information. Notably, each stage focuses on different attribute information. To achieve this, we designed an ESM that is applied to the $F_{i} \in \mathbb{R}^{B \times C \times N}$ features, allowing each point in the point cloud to autonomously select the features it requires, where $i$ denotes $i^{th}$ stage. For convenience, we denote the process of selecting features from the point cloud as:

\begin{equation}
    F_{select}^{i}= ESM(F_{i}) ,
\end{equation}
Specifically, $ESM$ considers a point-selection map with the same shape as $F_{i}$. First, the map is initialized, and during training, the values in the point-selection map are optimized. The initial values follow a normal distribution map~$N\left(\mu, \sigma^{2}\right)$ with mean $\mu=0$ and standard deviation $\sigma=0.01$, which provides the network with a moderate amount of perturbation to gradually optimize the parameters in the point-selection map within the [0, 1] range. To enable the point cloud to select information based on the point-selection map, it selects attributes according to the varying values within the map. Then, map is used as a mask, and multiplied element-wise with the input features $F_{i}$ to yield the final output $F_{select}^{i}$:
\begin{equation}
    F_{select}^{i}=F_{i}\cdot map
\end{equation}

Meantime, we obtain the feature representation of each sample in a common embedding space by projecting the feature $F_{i}$ to a common dimension represented by:
\begin{equation}
    F_{{i}}^{\prime}={MLP}(F_{i})
\end{equation}

Next, each point selects effective semantic features to compute its deformation of scaling and translation $(S,T)$, give by:
\begin{equation}
    (S,T)=\phi(F_{select}^i) , 
\end{equation}
where $\phi$ represents MLPs, $S$ denoted the scaling factor and $T$ denoted the translation.

Finally, during the decoding process into 3D space, this information is used to guide the corresponding features. 
\begin{equation}
    F_{i+1}=F_{i}^{\prime} \cdot S + T
\end{equation}

\begin{table*}[ht]
\centering
\resizebox{\textwidth}{!}{
\begin{tabular}{c||ccccc|ccccc}
\hline
 & \multicolumn{5}{c|}{CD } & \multicolumn{5}{c}{F-score@1\%  } \\
\hline
Methods & Point-E\cite{pointe} & 3DAttriFlow \cite{3dAttriflow} & PC²\cite{pc2} & BDM-B \cite{bdm} & \textbf{Ours}  & Point-E \cite{pointe} & 3DAttriFlow \cite{3dAttriflow} & PC² \cite{pc2}& BDM-B \cite{bdm}& \textbf{Ours} \\
 & (Arxiv 22) & (CVPR 22) & (CVPR 23) & (CVPR 24) &  & (Arxiv 22) & (CVPR 22) & (CVPR 23) & (CVPR 24) &  \\
\hline
Airplane & 17.44 & 2.11 & 4.12 & 4.01 & \textbf{1.91}  & 0.486 & 0.983 & 0.920 & 0.928 & \textbf{0.987} \\
Bench & 32.00 & 2.71 & 4.18 & 3.74 & \textbf{2.41}  & 0.188 & 0.978 & 0.921 & 0.930 & \textbf{0.985} \\
Cabinet & 22.59 & 2.66 & 6.08 & 6.03 & \textbf{2.49}  & 0.280 & 0.984 & 0.839 & 0.842 & \textbf{0.986} \\
Car & 23.55 & 2.50 & 4.74 & 4.82 & \textbf{2.36}  & 0.292 & 0.990 & 0.929 & 0.926 & \textbf{0.993} \\
Chair & 16.89 & 3.33 & 5.25 & 5.15 & \textbf{3.02}  & 0.374 & 0.966 & 0.883 & 0.888 & \textbf{0.976} \\
Display & 24.27 & 3.60 & 6.24 & 6.03 & \textbf{3.15}  & 0.300 & 0.952 & 0.832 & 0.838 & \textbf{0.971} \\
Lamp & 27.06 & 4.55 & 8.12 & 8.12 & \textbf{4.26}  & 0.290 & 0.898 & 0.705 & 0.711 & \textbf{0.907} \\
Loudspeaker & 20.33 & 4.16 & 7.55 & 7.35 & \textbf{3.72}  & 0.340 & 0.928 & 0.744 & 0.758 & \textbf{0.943} \\
Rifle & 15.90 & 1.94 & 2.70 & 2.71 & \textbf{1.77} & 0.498 & 0.986 & 0.964 & 0.966 & \textbf{0.990} \\
Sofa & 26.26 & 3.24 & 6.57 & 6.48 & \textbf{3.00} & 0.249 & 0.970 & 0.840 & 0.844 & \textbf{0.975} \\
Table & 25.38 & 2.85 & 5.91 & 5.83 & \textbf{2.67}  & 0.239 & 0.966 & 0.843 & 0.843 & \textbf{0.985} \\
Phone & 28.46 & 2.66 & 4.30 & 4.37 & \textbf{2.38} & 0.239 & 0.976 & 0.943 & 0.940 & \textbf{0.987} \\
Vessel & 18.02 & 2.96 & 4.36 & 4.28 & \textbf{2.70}  & 0.439 & 0.973 & 0.920 & 0.924 & \textbf{0.982} \\
\hline
Average & 22.93 & 3.02 & 5.39 & 5.30 & \textbf{2.76} & 0.324 & 0.965 & 0.868 & 0.872 & \textbf{0.974} \\
\hline
\end{tabular}
}
\caption{2D-to-3D reconstruction on ShapeNet dataset in terms of per-point L1 CD $\times10^{2}$ and F-score@1\%.}
\label{shapenet_cdfs}
\end{table*}

The benefits of this approach are threefold: first, it reduces feature entanglement by allowing each point to compute its positive and negative features from the multiple channels of the fused features, thereby effectively extracting the semantic cues each point needs; second, it reduces the computational load by excluding marginalized features that are less relevant from the computations; and third, it reduces inference time, as demonstrated by our comparison of the inference time of diffusion-based methods with our approach. This method not only streamlines the computational process but also enhances the precision and efficiency of the 3D reconstruction task.

\subsection{Training loss}
\label{sec:loss}
For 3D text encoder pre-training, as shown in Fig. \ref{pipeline}, for an object i, features $T_{{emb}}^{i}$ and $P_{{emb}}^{i}$ are extracted from learnable text prompt and 3D point cloud encoders. Then contrastive loss among each pair of modalities is computed as follows,
\begin{equation}
\label{multi-loss}
    \begin{aligned}
        L_{(T,P)} &= \sum_{(i,j)} -\frac{1}{2} \log \frac{\exp\left(\frac{{T_{{emb}}^{i} P_{{emb}}^{j}}}{\tau}\right)}{\sum_{k} \exp\left(\frac{{T_{{emb}}^{i} P_{{emb}}^{k}}}{\tau}\right)} \\
        &\quad - \frac{1}{2} \log \frac{\exp\left(\frac{{T_{{emb}}^{i} P_{{emb}}^{j}}}{\tau}\right)}{\sum_{k} \exp\left(\frac{{T_{{emb}}^{k} P_{{emb}}^{j}}}{\tau}\right)}.
    \end{aligned}
\end{equation}

where $T$ and $P$ represent two modalities and $(i, j)$ indicates a positive pair in each training batch. We use a learnable temperature parameter \(\tau\) as well, similar to CLIP \cite{clip}.

We train our MESC-3D model fully supervised using the Chamfer Distance(CD) loss. The CD measures the distance of each point to the other set:
\begin{equation}
\label{cd}
    \begin{aligned}\mathcal{L}_{\mathrm{CD}}(\mathcal{P}^{\mathrm{o}},\mathcal{P}^{\mathrm{t}})=&\frac{1}{2N}\sum_{p^{\mathrm{o}}\in\mathcal{P}^{\mathrm{o}}}\min_{p^{\mathrm{t}}\in\mathcal{P}^{\mathrm{t}}}\left\|p^{\mathrm{o}}-p^{\mathrm{t}}\right\|_{2}\\&+\frac{1}{2N}\sum_{p^{\mathrm{t}}\in\mathcal{P}^{\mathrm{t}}}\min_{p^{\mathrm{o}}\in\mathcal{P}^{\mathrm{o}}}\left\|p^{\mathrm{t}}-p^{\mathrm{o}}\right\|_{2}\end{aligned},
\end{equation}
where $\mathcal{P}^{\mathrm{o}}$ and $\mathcal{P}^{\mathrm{t}}$ represent the predicted point cloud and the corresponding ground truth point cloud respectively.

\section{Experiments}
% In this section, we present extensive experimental evaluations of MESC-3D on the ShapeNet and Pix3D datasets. We first describe the datasets and evaluation protocols. Next, we demonstrate the implementation details of the proposed methods. Finally, we report experimental evalutions of the proposed methods against state-of-the-art methods.
In this section, we experimentally evaluate the effectiveness of MESC-3D in 3D reconstruction task, and analyze the quantitative and qualitative results. The ablation studies will focus on effectiveness of each part of MESC-3D, visually analyze the generalization ability on base classes. To demonstrate the benefits of 3DSPL, we conduct experiments on the zero-shot 3D reconstruction task, where the model is evaluated on previously unseen classes.
\subsection{Datasets And Evalution Metrics}
Following \cite{3dAttriflow,pc2,bdm} and \cite{cheng2023sdfusion,li20233dqd}, we evaluate our method on the synthetic dataset ShapeNet \cite{chang2015shapenet} and the real-wolrd dataset Pix3D \cite{sun2018pix3d}.

\noindent\textbf{ShapeNet.} The ShapeNet Dataset is a collection of 3D CAD models corresponding to categories in the WordNet lexical database. We use a subset of the ShapeNet dataset consisting of 44K models and 13 major categories following the experimental setup used in 3DAttriflow \cite{3dAttriflow}. More specifically, renderings of each 3D model contains 24 random views of size 137 × 137. 

\noindent\textbf{Pix3D.} The Pix3D dataset is a large-scale single-image 3D shape modeling dataset that features pixel-level 2D-3D alignments. The dataset contains 395 3D models of nine object classes such as chairs, tables, and sofas.

\noindent\textbf{Evalution Metrics.} To evaluate the reconstruction quality of the proposed methods, we calculate Chamfer Distance (CD) between points clouds uniformly sampled from the ground truth and our prediction to measure the surface accuracy. We use Chamfer disatance described by Eq.~\ref{cd} to evaluate the performance of 3D reconstruction results. In addition, following PC² \cite{pc2}, we also take F-score@1\% as an extra metric. 
% \begin{equation}
%     \text{F-Score}(d)=\frac{2P(d)R(d)}{P(d)+R(d)}
% \end{equation}
For CD, the smaller is better. For F-score@1\%, the larger is better.

\begin{figure*}[!ht]
    \centering
    \setlength{\abovecaptionskip}{0.2cm}
    \includegraphics[width=1\linewidth]{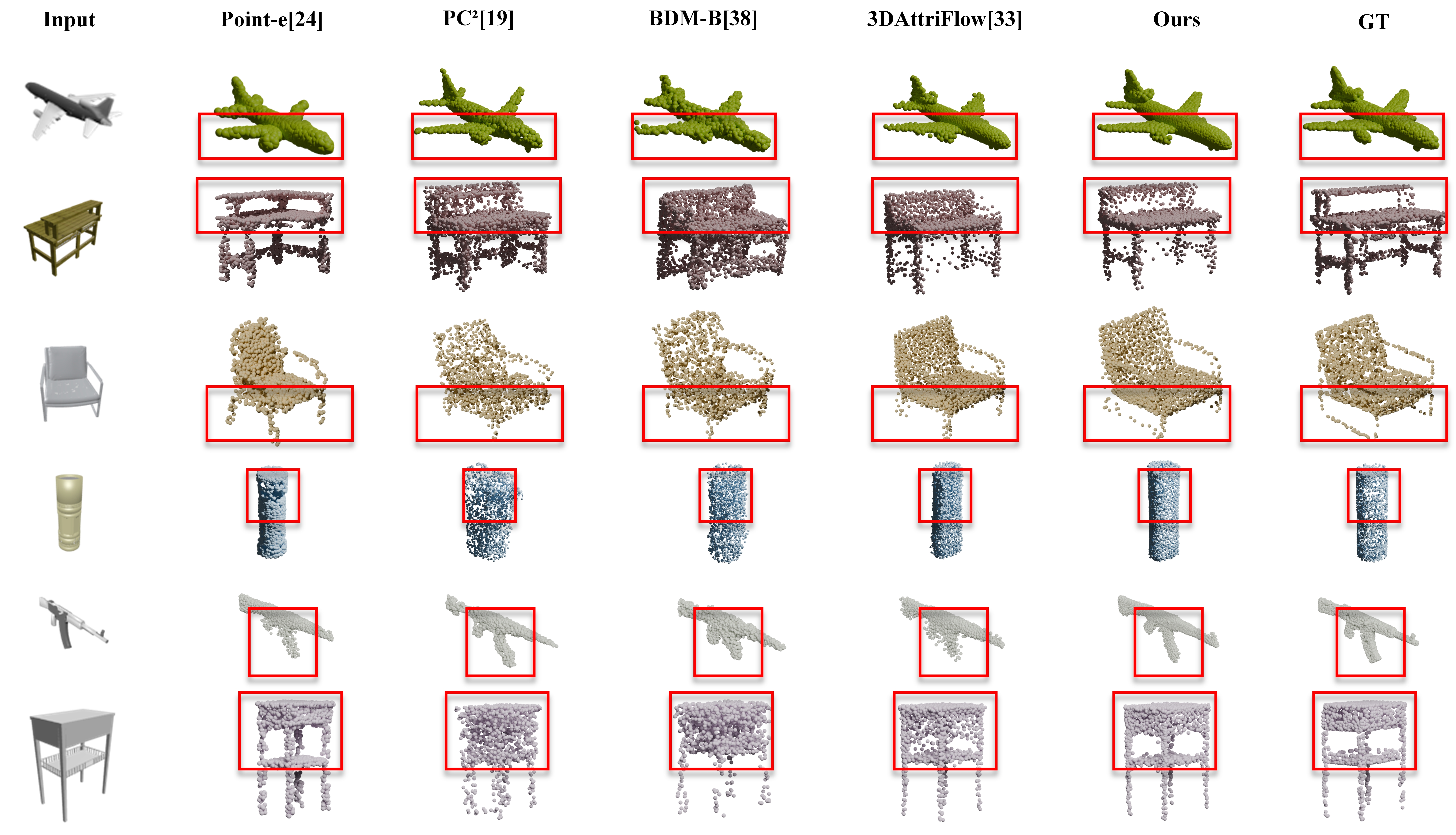} 
    \caption{Visual comparison of 2D-to-3D reconstruction results with different methods under ShapeNet dataset. Additional qualitative results are provided in the supplementary material.}
    \vspace{0.5cm}
    \label{shapenet_vision}
\end{figure*}

\subsection{Implementation Details}
\textbf{Pre-training.} For 3DSPL, we use the point cloud data from the ground truth of our training and validation sets, without utilizing the ground truth of the test set for training. We freeze the point cloud encoder and only update the text encoder's parameters during pre-training. MESC-3D is trained for 200 epochs. We use 64 as the batch size, $10^{-4}$ as the learning rate, and Adam as the optimizer.

\noindent\textbf{3D Reconstruction.} On ShapeNet, we use the learning rate of $10^{-3}$ and Adam as the optimizer.The initial learning rate is set to $10^{-3}$ and gradually decays to $10^{-5}$. We train for 400 epochs with batch size 24 on an RTX 3090 GPU. 

\subsection{Comparison with SOTAs}
We benchmark our method against SOTA approaches in point cloud reconstruction, including those based on traditional deep learning \cite{3dr2n2,psgn,pixel2mesh,AtlasNet,OCcnet,3dAttriflow} and diffusion methods \cite{pointe,pc2,bdm,li20233dqd}.

\noindent\textbf{Quantitative Comparison.} We present results for 13 categories in the widely-used ShapeNet benchmark. Tab. \ref{shapenet_cdfs} shows a quantitative comparison between our method and state-of-the-art point cloud reconstruction methods. Our method consistently outperforms the SOTA method across all 13 classes and on average, demonstrating its capability to achieve superior performance comparable to prior methods. Through the improvement in metrics across various categories, we see that our method performs better on both simple and complex objects, such as ``display” and ``airplane”. Overall, our approach achieves an average Chamfer Distance of 2.76 and an F-score@1\% of 0.974, outperforming the best competing method by a notable margin.

We also evaluate on the Pix3D\cite{sun2018pix3d} dataset in Tab. \ref{pix3d_biaoge}. It can be seen that our method remarkably improves the performance and achieves state-of-the-art. On average, we achieve 1.37 CD, while other diffusion methods have around 7, showing that our approach attains a much better overall object reconstruction quality.

\begin{table}[!h]
\centering
\vspace{0.35cm}
\caption{2D-to-3D reconstruction on Pix3D dataset in terms of per-point L1 CD $\times10^{2}$ and F-score@1\%.}
\fontsize{5}{7}\selectfont
\resizebox{\linewidth}{!}{
\begin{tabular}{ccc}
\hline
Methods  & CD  & F-score@1\%   \\
\hline
SDFusion(CVPR'23) \cite{cheng2023sdfusion} & 7.22 & 0.772    \\
3DQD(CVPR'23) \cite{li20233dqd}    & 8.65 & 0.693    \\                           
\textbf{Ours}     & \textbf{1.37} & \textbf{0.990}   \\     
\hline
\end{tabular}}
\label{pix3d_biaoge}
\end{table}
\vspace{0.35cm}

\begin{table*}[ht]
\centering
\setlength{\abovecaptionskip}{0.1cm}
\caption{Impact of different modules in our MESC-3D in terms of L2 CD$\times10^{3}$. We show the results of integrating different modules into the baseline. ESM denoted as Effective Semantic Mining Module and 3DSPL denoted as 3D Semantic Prior Learning Module. }
\label{shapenet_cd2}
\resizebox{\textwidth}{!}{
\begin{tabular}{cccccccccccccccc}
\toprule
Model & Avg-CD  & Airplane & Bench & Cabinet & Car & Chair & Display & Lamp & Loud. & Rifle & Sofa & Table & Tele. & Vessel \\
\midrule
Baseline  & 5.68    & 2.22     & 4.14  & 3.59    & 2.21 & 4.83  & 8.66    & 14.61 & 12.66 & 1.87  & 5.74  & 4.48  & 4.89  & 3.88   \\
Baseline + ESM & 3.69    & 1.79     & 2.68  & 2.62    & 1.87 & 3.36  & 4.85    & 10.20 & 7.17  & 1.62  & 3.45  & 3.29  & 2.29  & 2.74   \\
Baseline + ESM + 3DSPL & \textbf{3.20} & \textbf{1.65} & \textbf{2.52} & \textbf{2.36} & \textbf{1.83} & \textbf{3.24} & \textbf{3.58} & \textbf{8.45} & \textbf{5.32} & \textbf{1.53} & \textbf{3.32} & \textbf{3.06} & \textbf{2.18} & \textbf{2.63} \\
\bottomrule
\end{tabular}}
\vspace{0.35cm}
\end{table*}

\noindent\textbf{Qualitative Comparison.} To provide a more comprehensive comparison with SOTAs, We also presents the qualitative results as seen in the Fig. \ref{shapenet_vision}. It is evident that 3DAttriFlow struggles to model regions with high uncertainty from a single view, such as the two ram air turbines on the airplane and the backrest of the bench, which are all occluded in this perspective and lack sufficient information. In contrast, by incorporating 3D prior, our method effectively reconstructs these occluded areas. Additionally, BDM-B produces very blurred reconstructions for some complex objects, such as the shape of the chair and the magazine of the rifle. Relative to prior methods, our approach generates shapes with significantly finer levels of detail.

\begin{figure}[!h]
    \centering
    \includegraphics[width=1\linewidth]{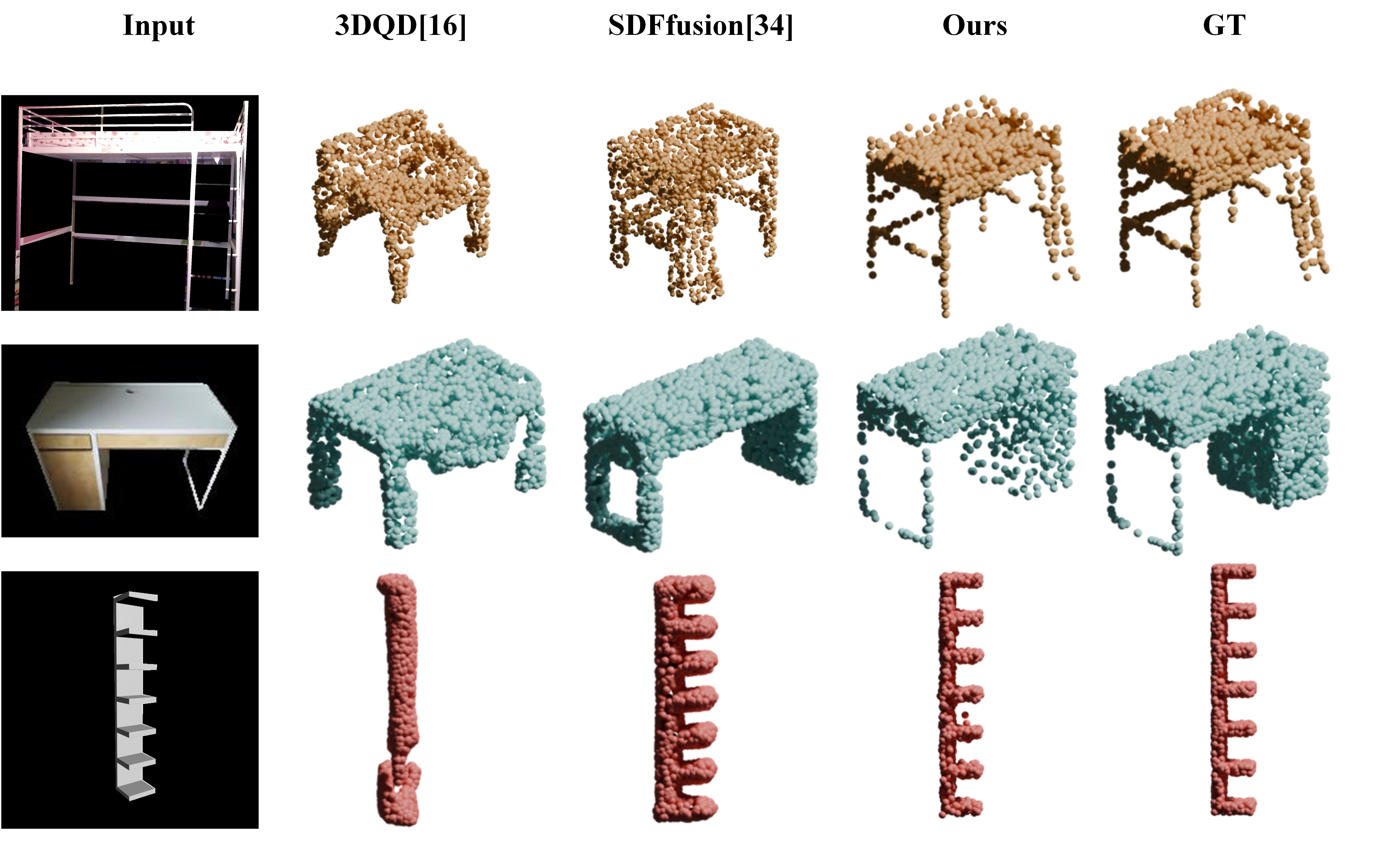} 
    \caption{Visual comparison of 2D-to-3D reconstruction results with different methods under Pix3D dataset.}
    \label{pix3d_vision}
\end{figure}

Fig. \ref{pix3d_vision} presents qualitative results on Pix3D dataset. Our method trained on real-world data recovers not only realistic geometry, but also fine-grained, such as the thickness of the bookcase and the components of the bed with a ladder. This further demonstrates that its better mining effective semantic cues to guide 3D reconstruction.  
\vspace{0.5cm}
\begin{table}[h]
\setlength\tabcolsep{12pt}
\centering
\setlength{\abovecaptionskip}{0.1cm}
\vspace{0.35cm}
\caption{Abaltion Study on Image Encoder. We present results in terms of L2 CD$\times10^{3}$, using a frozen CLIP model, a fine-tuned CLIP model, and a trainable ResNet18 network architecture on vessel.}
\label{image_encoder}
\begin{tabular}{clc}
\toprule
Image Encoder     & CD  & F-score@1\%  \\
\midrule
CLIP(Pre-trained) & 4.71 & 0.954                                    \\
CLIP(Fine-tuned)  & 4.24 & 0.966                                    \\
\textbf{Resnet18(Ours) }         & \textbf{2.74} &  \textbf{0.980}                                   \\
\bottomrule
\end{tabular}
\end{table}
\subsection{Ablation Study}
\begin{figure}[!h]
    \centering
    \includegraphics[width=\linewidth]{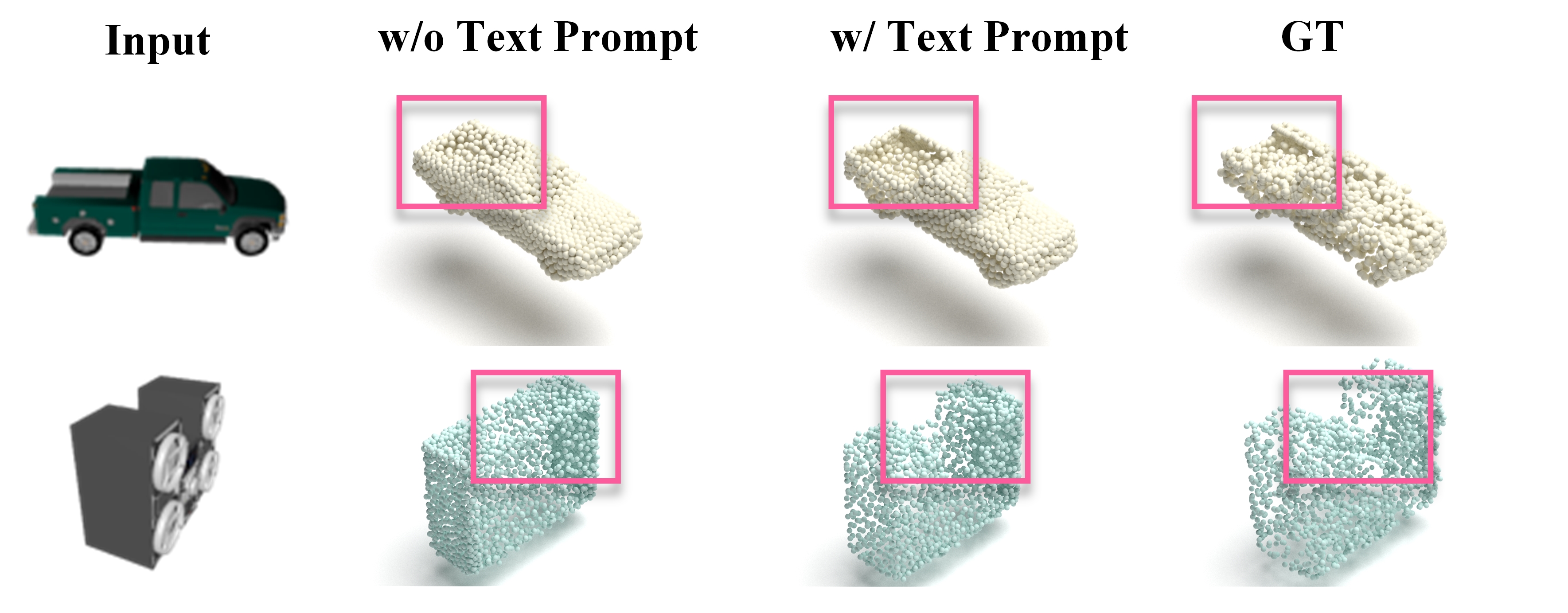} 
    \caption{Ablation Study on learnable text prompt. Visual results on ShapeNet.}
    \label{ablation_vision}
\end{figure}

\textbf{Abaltion Study on Model Architecture.} Tab. \ref{shapenet_cd2} shows the impact of integrating different components into the baseline. The baseline adopts original Multimodal Interlaced Transformer (MIT) and 3D Decoder to predict the point cloud. When the ESM was introduced, the score improved from 5.68 to 3.69, outperforming the baseline by 35\%. Furthermore, when the prior was introduced through the 3DSPL, the occluded areas in the image have been reconstructed with enhanced details, as seen in Fig. \ref{ablation_vision}. The results validate that each component contributes to the performance of our method.

\noindent\textbf{Abaltion Study on Image Encoder.} We conduct ablation study on the backbone without the introduction of text prompt, and the evaluation metrics were measured only on the vessel. Tab. \ref{image_encoder} presents the ablation study on the pre-training and fine-tuning of the image encoder. In this ablation, our model incorporates only the ESM, and the experimental results are measured exclusively on the vessel. 
% We compare the quantitative results using Chamfer Distance ($CD\ell_2$) and F-score@1\%.
\vspace{0.35cm}

\begin{table}[!h]
\centering
\setlength{\abovecaptionskip}{0.1cm}
\vspace{0.2cm}
\caption{Ablation Study on Learnable Text Prompt Training Strategy. we evaluated them by L2 CD$\times10^{3}$  metrics on vessel.}
\label{s1_strategy}
\small
\resizebox{\linewidth}{!}{
\begin{tabular}{cccc}
\toprule
Text Encoder         & PC Encoder              & CD  & F-score@1\%  \\
\midrule
CLIP (Pre-trained)      & ULIP             & 2.80 & 0.978                                    \\
CLIP (Pre-trained)      & PointMAE         & 2.74 & 0.980                                    \\
\textbf{CLIP (DPT)(Ours)}           & \textbf{ULIP}             &  \textbf{2.63} &  \textbf{0.982}                 \\     
\bottomrule
\end{tabular}}
\end{table}

\noindent\textbf{Ablation Study on Learnable Text Prompt Training Strategy.}
Tab. \ref{s1_strategy} shows different training strategies for the first-stage learnable text prompt. We employed various Text Encoders, PC Encoders, and Image Encoders. In the first two strategies, we used pre-trained encoders and applied learnable text prompt with point cloud and image triad alignment strategies. The L2 CD scores for these strategies are similar. In the final strategy, we introduced DPT to fine-tune the Text Encoder, combined with learnable text prompt and point cloud pair alignment strategies. The experiments demonstrate that incorporating DPT leads to a better learning of 3D geometric prior.

\noindent\textbf{Ablation Study on Introduce Manner of Learnable Text Prompt.}
Tab. \ref{learnable_prompt} shows different methods for incorporating 3D prior into the point cloud reconstruction stage, including addition, fusion, and concatenation of point cloud and image fusion features. The experimental results indicate that the three-modal fusion strategy performs the worst. Directly adding the 3D prior to the fused features also did not significantly improve performance. However, concatenating the 3D prior with the fused features led to an improvement in the L2 CD score to 2.80. This method's advantage lies in preserving the original 3D prior, thereby providing more intuitive feature selection for point cloud reconstruction.

\begin{table}[!h]
\centering
\setlength{\abovecaptionskip}{0.1cm}
\setlength\tabcolsep{8pt}
\caption{Introduce of Learnable Promp. 
Here, \textcircled{\scalebox{0.7}M} represents the MultiModal Interlace Transformer, {\textcircled{\scalebox{0.8}C}} stands for the concatenation operation between two vectors, and \textcircled{+} denotes element-wise addition between two vectors. The following operations are performed sequentially from left to right. We evaluated them on vessel in terms of L2 CD$\times10^{3}$.}
\label{learnable_prompt}
\begin{tabular}{ccc}
\toprule
Methods    & CD  & F-score@1\%  \\
\midrule
$F_{sem}$ \textcircled{\scalebox{0.7}M} $T_{emb}$ \textcircled{\scalebox{0.7}M} $F_{geo}$ & 3.26 & 0.974                                    \\
$F_{sem}$ \textcircled{\scalebox{0.7}M} $F_{geo}$ \textcircled{+} $T_{emb}$                & 2.98 & 0.976                                    \\
$F_{sem}$ \textcircled{\scalebox{0.8}C} $T_{emb}$ \textcircled{\scalebox{0.7}M} $F_{geo}$        & 2.92 & 0.977                                    \\
$F_{sem}$ \textcircled{\scalebox{0.7}M} $F_{geo}$ 
 \textcircled{\scalebox{0.8}C} $T_{emb}$        & \textbf{2.80} &  \textbf{0.978}         \\   
\bottomrule
\end{tabular}
\end{table}

\begin{table}[!h]
\centering
\setlength{\abovecaptionskip}{0.1cm}
\caption{Hyperparameter Analysis. Here, $H$ represents the number of attention heads, $D$ denotes the number of attention layers, and Group refers to the grouping strategy for the 768 feature channels. Specifically, $Group1$ divides the feature channels into 24 groups, each containing 32 channels, resulting in a grouping of (24,32); whereas $Group2$ denoted as a grouping of (96,8).}
\label{param}
\resizebox{\linewidth}{!}{
\begin{tabular}{cccccc}
\toprule
$Head$ & $Depth$ & $Group1$ & $Group2$ &$CD\ell_2$ & $F-score@1\%$\\
\midrule
4    & 4     & \checkmark & $\times$  & 3.05  &  0.975                                       \\
4    & 4     & $\times$& \checkmark & 3.06  & 0.975                                       \\
8    & 4     & \checkmark & $\times$ & \textbf{2.74}& \textbf{0.980}                   \\
8    & 4     & $\times$& \checkmark & 2.94 & 0.976                                    \\
8    & 8     & \checkmark & $\times$ & 2.98 & 0.976                                    \\
8    & 12    &  $\times$ & \checkmark & 2.86 & 0.978                                    \\
8    & 16    &  $\times$ & \checkmark & 2.88 & 0.977                              \\\bottomrule
\end{tabular}}
\vspace{0.35cm}
\end{table}
\noindent\textbf{Hyperparameter Analysis.}
We experimented with 7 different combinations of MIT parameter as illustrated in Tab. \ref{param}. The results indicate that the optimal parameter combination is ($H$=4, $D$=4, $Group1$).
\subsection{Further Analysis}
\begin{figure}[!h]
    \centering
    \includegraphics[width=1\linewidth]{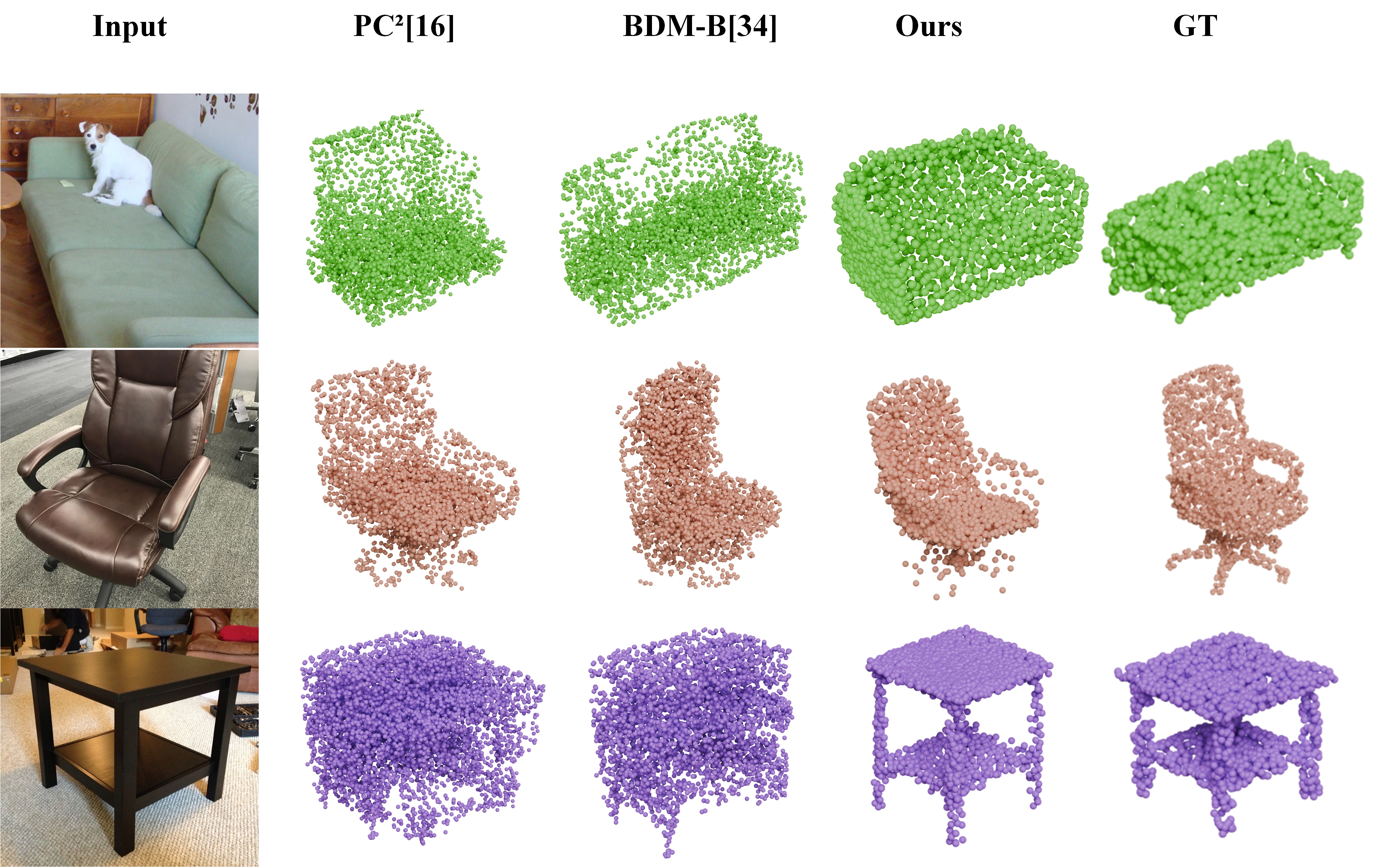} 
    \caption{Generalization on base classes with various methods.}
    \vspace{0.35cm}
    \label{pix3d_ablation}
\end{figure}

\noindent\textbf{Generalization on Base Classes.}
We use ShapeNet as the training set and Pix3D as the test set, keeping other settings consistent and without introducing prompt learning. By visually comparing the model predictions with the Ground Truth, as seen in Fig. \ref{pix3d_ablation}, we observe that our model successfully reconstructs the square-shaped tabletop and the four table legs. This result demonstrates the model’s effective capability in mining semantic features and its strong generalization ability to unseen basic classes.

\begin{figure}[!h]
    \centering
    \includegraphics[width=1\linewidth]{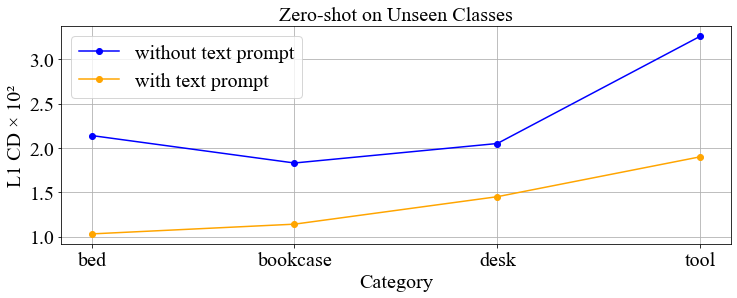} 
    \caption{Zero-shot on Unseen Classes in Pix3D dataset.}
    \vspace{0.35cm}
    \label{zero-shot-ablation}
\end{figure}
\noindent\textbf{Zero-shot on Unseen Classes.}
We conducted a quantitative analysis of zero-shot performance through an ablation study on learnable text prompt, where the text prompt were trained only on the 13 categories (excluding bed, bookcase, desk, and tool). As shown in Fig. \ref{zero-shot-ablation}, introducing the text prompt led to a significant improvement in the metrics. By incorporating 3D prior, the reconstruction of unseen classes was effectively aided, thereby demonstrating the zero-shot capability enabled by learnable text prompt.

\section{Conclusion}
In this work, we propose a novel 3D point cloud reconstruction method from a single-view image by mining effective semantic cues. To ensure precision and efficiency in reconstruction, we designed two key components to achieve competitive results on the reconstruction task. The first is the ESM, which enables the point cloud to autonomously select relevant semantic information during the decoding process, overcoming the issue of deeply entangled with each other. The second component is the 3DSPL, which introduces learnable text prompts and contrastive learning to emulate the human-like ability to learn 3D prior knowledge. To demonstrate the effectiveness of our method, we conduct extensive evaluations on both synthetic and real-world datasets, achieving performance that surpasses previous state-of-the-art methods. Furthermore, our work explores generalization to base classes and zero-shot capability on unseen classes, which may provide inspiration for future research.

{
    \small
    \bibliographystyle{ieeenat_fullname}
    \bibliography{main}
}

\newpage
\leftline{\Large{\textbf{Supplementary Material}}}
\renewcommand\thesection{\Alph{section}}
\setcounter{section}{0}
\vspace{0.5cm}
\noindent To provide a more comprehensive explanation of our method, this supplementary material includes detailed information on various aspects of our method:
\begin{itemize}
    \item Training and Dataset Details
    \item Complexity Analysis
    \item More Visualization Results
    \item Algorithm MSEC-3D Explanation
\end{itemize}

\section{More Implementation Details}

\textbf{Training for Two Stages. }For the 3D input, we follow the experimental settings of 3DAttriflow, uniformly sampling $N_{p}$ = 2048. Our learnable text prompts and 3D reconstruction are trained in two separate stages, with both stages using only data from the training and validation sets, excluding any data from the test set. It is worth noting that our approach adopts a all-categories strategy., unlike methods such as PC² and BDM, which rely on single-category training for diffusion.

\noindent\textbf{Generalization Capability Experiments. }For the generalization capability experiment , we replace the image encoder ResNet18 with a CLIP large model and DPT, fine-tuned on ShapeNet but tested on Pix3D. Finally, we conduct robustness testing on Pix3D, demonstrating that our network effectively mining semantic information for 3D shape reconstruction.

\noindent\textbf{Zero-shot Capability Experiments. }The learnable text prompt is trained exclusively on the ShapeNet dataset (comprising 13 categories), and subsequently embedded to provide prior guidance for the reconstruction of previously unseen categories.
\section{Dataset Details}
We continue testing qualitative results, parameter numbers, and inference time on a subset of ShapeNet. In robustness experiments, we not only test on Pix3D but also download some online photos for 3D reconstruction, further validating the robustness and efficiency of our network.

\section{Complexity Analysis}
As shown in Tab. \ref{complexity}, the comparison results indicate that the inference speed of diffusion models is significantly slower than ours, and they also use the most parameters. Compared to 3DAttriFlow, MESC3D performs on par with prior work. Although incorporating text prompt encoding naturally slows down inference slightly, our CDL2 metric greatly exceeds theirs. We also conducted an impact test on the number of point clouds as seen in Tab. \ref{number}. When increasing the number of point clouds from 2048 to 8192, the effect on our training and inference times was minimal.

\begin{table}[!h]
\caption{Complexity and inference time of different methods. w/o and w represent without and with text prompt respectively.}
\label{complexity}
\resizebox{\linewidth}{!}{
\begin{tabular}{c|cccl}
\toprule
Methods     & Params & Infer time & Avg-$CD\ell_2$ (×$10^{3}$) &  \\
\midrule
Point-e     & 80.94M & 55.215s    & 155           &  \\
3DAttriFlow & 20.92M & 0.117s     & 4.08          &  \\
PC² & 27.65M & 2.800s     & 5.39          &  \\
BDM-B & 49.71M & 7.602s     & 5.3          &  \\
Ours (w/o)   & 24.05M & 0.165s     & 3.69          &  \\
Ours (w)    & 24.97M & 0.548s     & 3.22 & \\
\bottomrule
\end{tabular}}
\end{table}

\begin{table}[!h]
\setlength\tabcolsep{30pt}
\centering
\caption{Impact of the number of point cloud on inference time.}
\label{number}
\begin{tabular}{c|c}
\toprule
Number of points     & Infer time \\
\midrule
Ours(w/o)2048                         & 0.165s                         \\
Ours(w/o)8192                         & 0.309s                         \\
\bottomrule
\end{tabular}
\end{table}

\begin{algorithm*}[t]
    \caption{MESC-3D:Mining Effective Semantic Cues for 3D Reconstruction from a Single Image}
    \label{alg:msec3d}
    \renewcommand{\algorithmicrequire}{\textbf{Input:}}
    \renewcommand{\algorithmicensure}{\textbf{Output:}}
    \begin{algorithmic}[1]
        \REQUIRE $I$ (image), $P$ (point cloud)
        \ENSURE $\mathbf{P}_{\text{pred}}$
        
        \STATE Extract image features: $\mathbf{I}_{\text{feat}} = \text{ResNet18}(I)$
        \STATE Extract point cloud features: $\mathbf{P}_{\text{feat}} = \text{PointMAE}(P)$
        
        \STATE Initialize $Q_0$ as random query values
        \FOR{each layer $t = 1$ to $T$}  
            \IF{t is even}
                \STATE Set $Q^t = \mathbf{P}_{\text{feat}}$, $K^t = V^t = \mathbf{I}_{\text{feat}}$
            \ELSE
                \STATE Set $Q^t = \mathbf{I}_{\text{feat}}$, $K^t = V^t = \mathbf{P}_{\text{feat}}$
            \ENDIF
            \STATE Perform attention: $\mathbf{F}_{\text{fusion}}^t = \text{Attention}(Q^t, K^t, V^t)$
            \STATE Update query: $Q^{t+1} = \mathbf{F}_{\text{fusion}}^t$
        \ENDFOR

        \STATE Initialize $\text{dec\_dim} = [768, 512, 256, 128, 64, 32]$
        \FOR{each layer $l = 1$ to $L$}  
            \STATE Compute downsampled features: \\ $\mathbf{F}_{\text{down}}^l = \text{conv}_l(\mathbf{F}_{\text{fusion}}^{l-1})$
            \STATE Select features: \\ $\mathbf{F}_{\text{select}}^l = \text{map}_l(\mathbf{F}_{\text{fusion}}^{l-1})$
            \STATE Normalize and fuse features: \\ $\mathbf{F}_{\text{next}}^l = \text{AdaptivePointNorm}(\mathbf{F}_{\text{down}}^l, \mathbf{F}_{\text{select}}^l)$
            \STATE Update: $\mathbf{F}_{\text{fusion}}^l = \mathbf{F}_{\text{next}}^l$
        \ENDFOR

        \STATE $\mathbf{F}_{\text{final}} = \mathbf{F}_{next}^{L}$  \COMMENT{Final fused features}

        \STATE \textbf{MLP for Point Cloud Reconstruction:}
        \STATE $\mathbf{P}_{\text{pred}} = \text{MLP}(\mathbf{F}_{\text{final}})$  \COMMENT{Apply MLP to map to point cloud features}
        \RETURN $\mathbf{P}_{\text{pred}}$ 
    \end{algorithmic}
\end{algorithm*}
\begin{figure*}[h]
    \centering
    \includegraphics[width=0.9\linewidth]{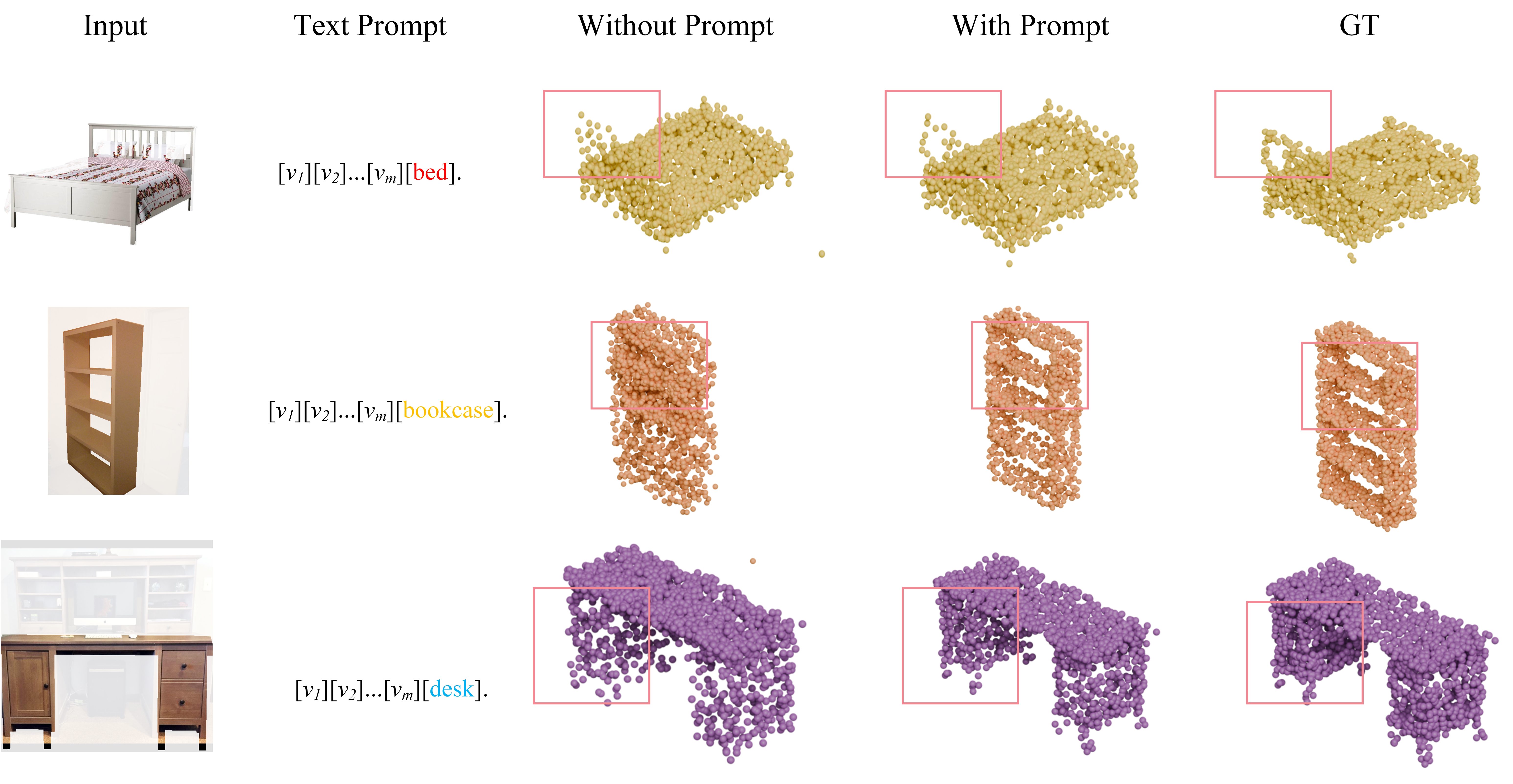}
    \caption{\textbf{Demonstration of the zero-shot ability of learnable text prompt, enabling detailed 3D shape reconstruction for unseen object categories.}}
    \label{fig:s_1}
\end{figure*}

\begin{figure*}[h]
    \centering
    \includegraphics[width=\linewidth]{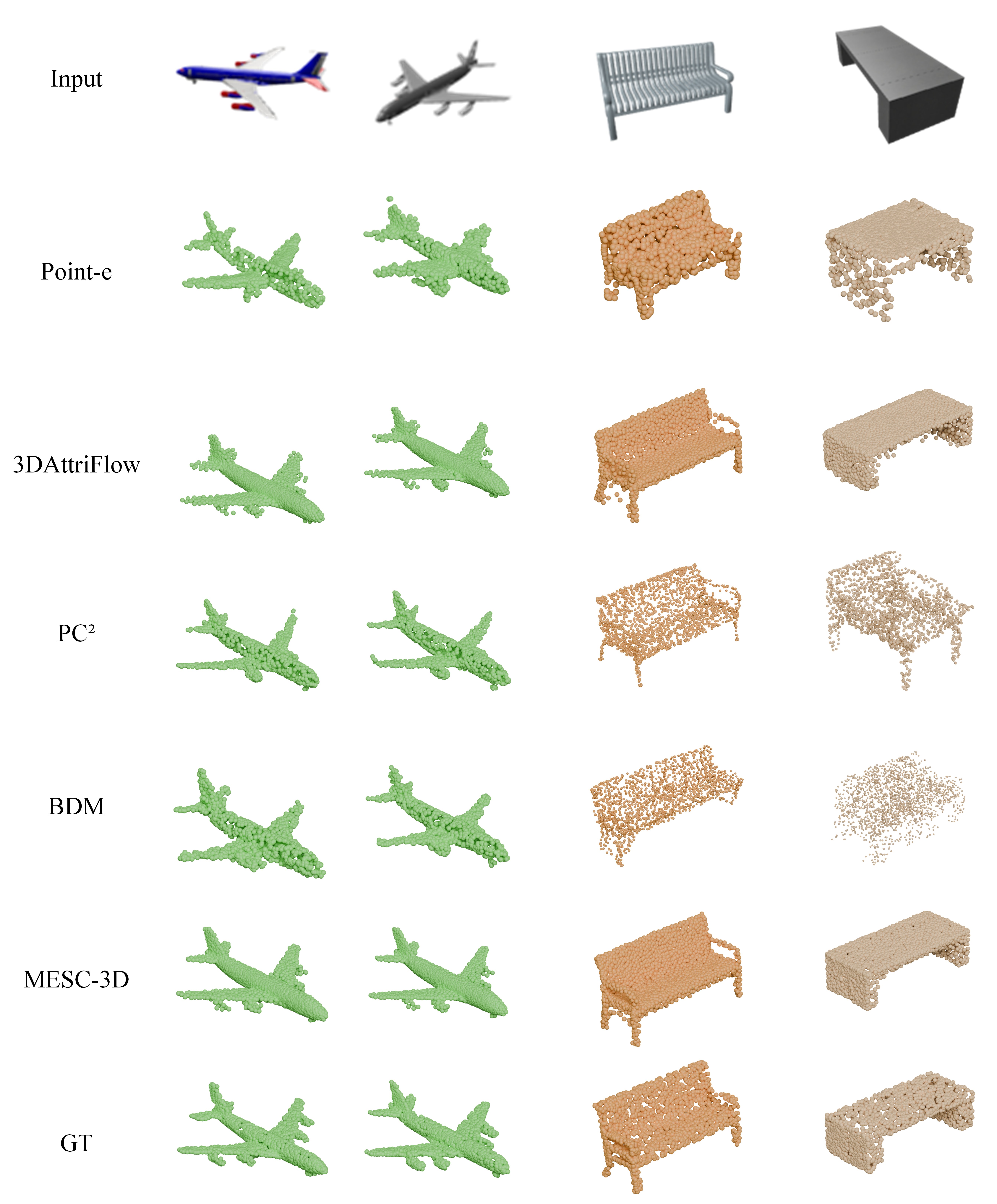}
    \caption{\textbf{Visual comparison of 2D-to-3D reconstruction results with different methods on ``airplane" and ``bench" in ShapeNet dataset.}}
    \label{fig:s0}
\end{figure*}

\section{More Visualization Results}
\begin{figure*}[h]
    \centering
    \includegraphics[width=\linewidth]{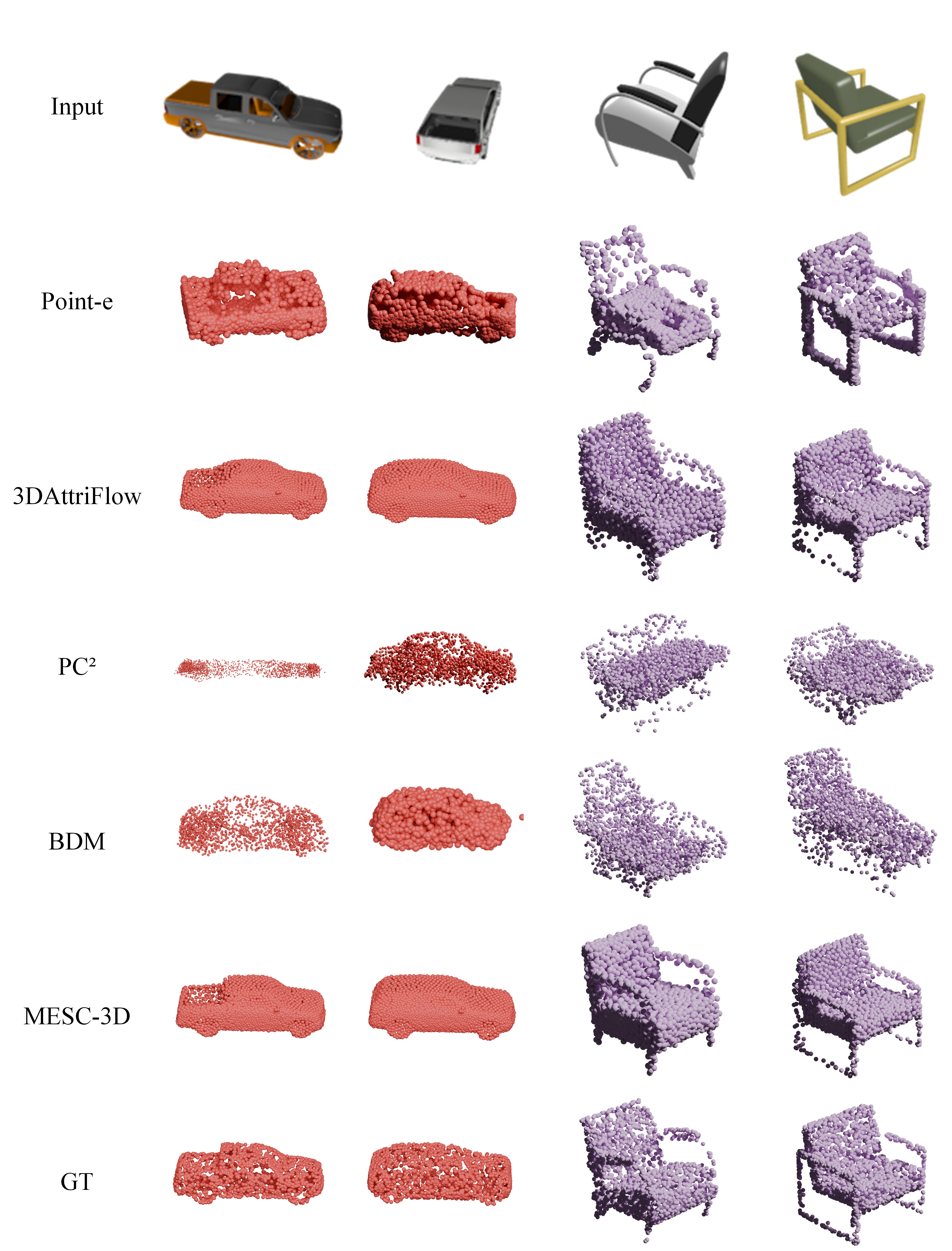}
    \caption{\textbf{Visual comparison of 2D-to-3D reconstruction results with different methods on ``car" and ``chair" in ShapeNet dataset.}}
    \label{fig:s1}
\end{figure*}

We offer additional visualization results on the ShapeNet dataset that demonstrate the superior performance of our method in recovering occluded regions from a single image. For example, our method successfully reconstructs the fully occluded sofa cushion as seen in Fig. \ref{fig:s1}, and the recovery of the truck bed is remarkable. Additionally, we excel in categories with objects that have fine details, such as the tail of the airplane and the shape recovery of the fighter jet as seen in Fig. \ref{fig:s0}. Compared to the diffusion-based method, our network has three main advantages:
\begin{itemize}
    \item Accurate foreground-background identification, ensuring the correct object is reconstructed from a single image with a higher reconstruction category accuracy. 
    \item Effective utilization of semantic information to guide the 3D reconstruction. 
    \item Consistency in results. Repeated inputs of the same image yield consistent output, while Point-E produces varied results each time.
\end{itemize}

Fig. \ref{fig:s_1} illustrates the zero-shot capability introduced by learnable text prompt. 

The detailed steps and implementation of the MESC-3D algorithm are provided in Algorithm~\ref{alg:msec3d}. In summary, our model demonstrates robust performance.

\thispagestyle{empty} % 去除页眉页脚

\begin{figure*}[h]
    \centering
    \includegraphics[width=\linewidth]{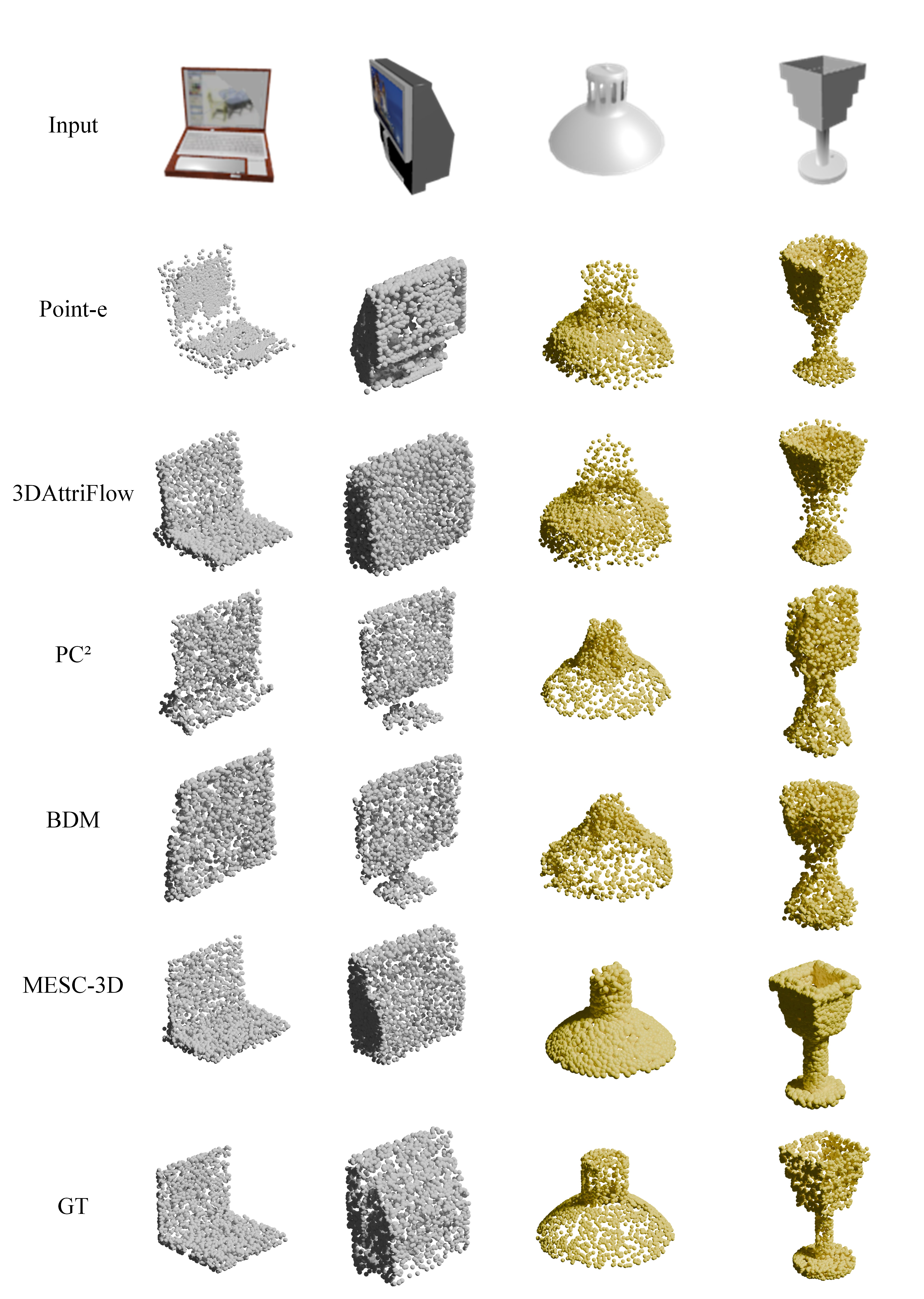}
    \caption{\textbf{Visual comparison of 2D-to-3D reconstruction results with different methods on ``display" and ``lamp" in ShapeNet dataset.}}
    \label{fig:s2}
\end{figure*}

\begin{figure*}[h]
    \centering
    \includegraphics[width=\textwidth]{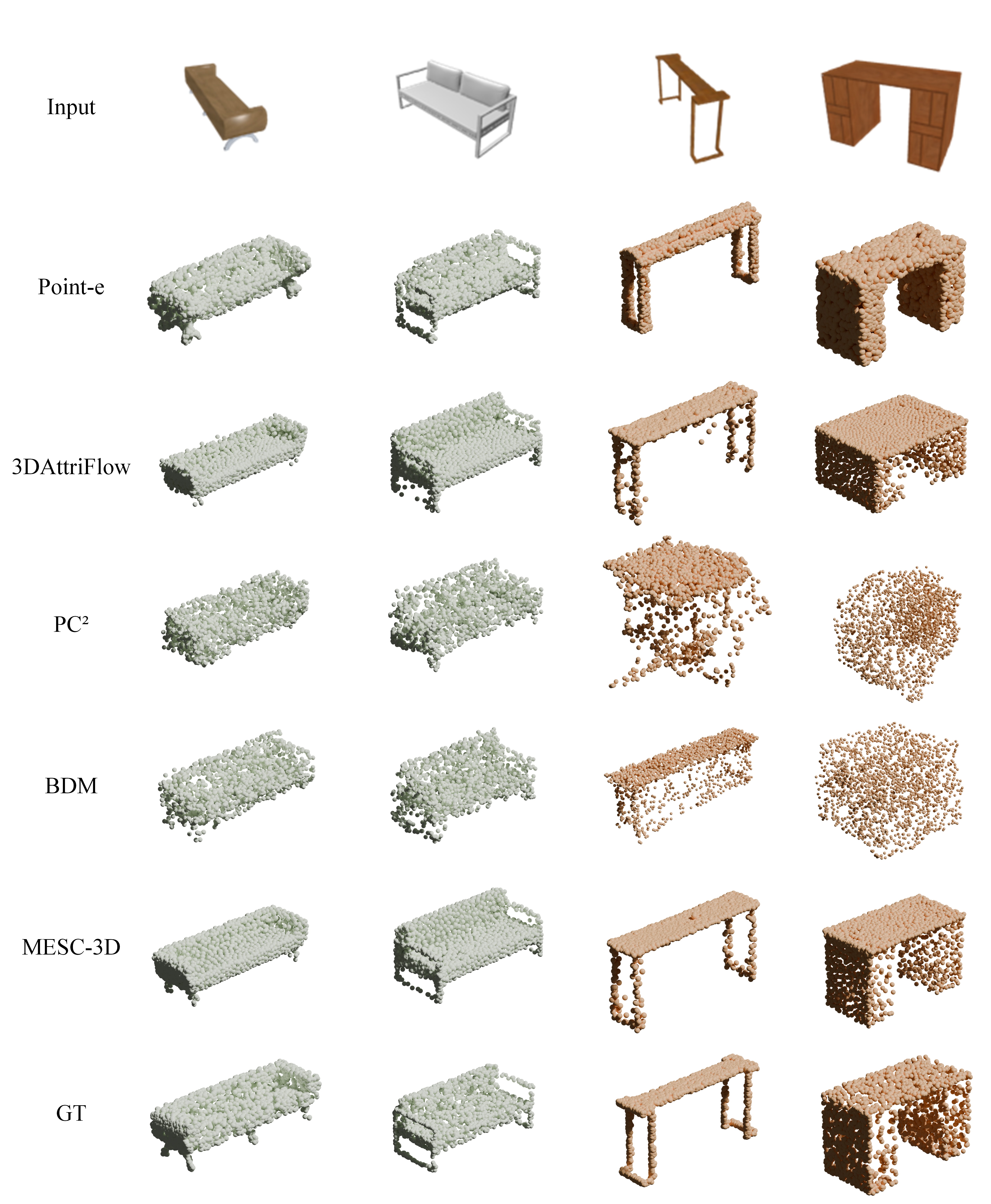}
    \caption{\textbf{Visual comparison of 2D-to-3D reconstruction results with different methods on ``sofa" and ``table" in ShapeNet dataset.}}
    \label{fig:s3}
\end{figure*}

\begin{figure*}[h]
    \centering
    \includegraphics[width=\linewidth]{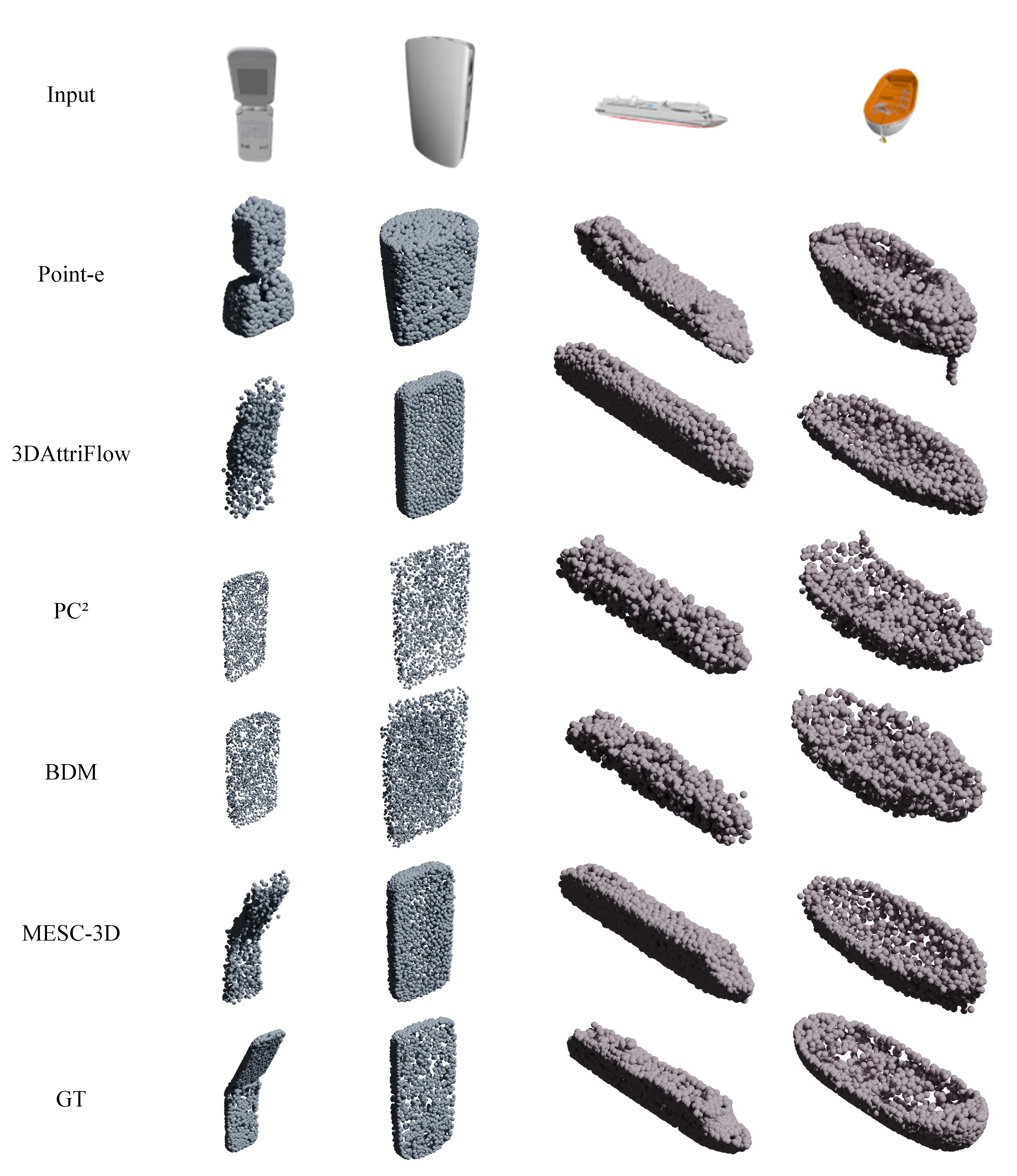}
    \caption{\textbf{Visual comparison of 2D-to-3D reconstruction results with different methods on ``telephone" and ``vessel" in ShapeNet dataset.}}
    \label{fig:s4}
\end{figure*}

\end{document}